\documentclass{article}

     \usepackage[final]{svrhm_2022}

\usepackage[utf8]{inputenc} 
\usepackage[T1]{fontenc}    
\usepackage{hyperref}       
\usepackage{url}            
\usepackage{booktabs}       
\usepackage{amsfonts}       
\usepackage{nicefrac}       
\usepackage{microtype}      
\usepackage{xcolor}         

\usepackage{graphicx}
\usepackage{bm}

\title{Sequence and Circle: Exploring the Relationship Between Patches}

\author{
    Zhengyang Yu \\
    Frankfurt Institute for Advanced Studies \\
    Xidian FIAS Joint Research Center \\
    Frankfurt am Main, Germany \\
    \texttt{zhyu@fias.uni-frankfurt.de} \\
    \And
    Jochen Triesch \\
    Frankfurt Institute for Advanced Studies \\
    Frankfurt am Main, Germany \\
    \texttt{triesch@fias.uni-frankfurt.de}
}

\begin{document}

\maketitle

\begin{abstract}
The vision transformer (ViT) has achieved state-of-the-art results in various vision tasks. It utilizes a learnable position embedding (PE) mechanism to encode the location of each image patch. However, it is presently unclear if this learnable PE is really necessary and what its benefits are. This paper explores two alternative ways of encoding the location of individual patches that exploit prior knowledge about their spatial arrangement. One is called the sequence relationship embedding (SRE), and the other is called the circle relationship embedding (CRE). Among them, the SRE considers all patches to be in order, and adjacent patches have the same interval distance. The CRE considers the central patch as the center of the circle and measures the distance of the remaining patches from the center based on the four neighborhoods principle. Multiple concentric circles with different radii combine different patches. Finally, we implemented these two relations on three classic ViTs and tested them on four popular datasets. Experiments show that SRE and CRE can replace PE to reduce the random learnable parameters while achieving the same performance. Combining SRE or CRE with PE gets better performance than only using PE.
\end{abstract}

\section{Introduction}

Vision Transformer (ViT) utilizes the self-attention mechanism and the way of building image patches to achieve global modeling capabilities while reducing huge computational costs \citep{dosovitskiy2020image,guo2022attention,han2022survey}. It retains the advantages of the transformer in NLP \citep{vaswani2017attention,devlin2018bert,wang2018glue}, catches global input relationships, parallelize computations, and achieve performance or surpass comparable to convolutional neural networks (CNN) on various computer vision tasks \citep{he2016deep, li2022exploring}, such as image classification \citep{bi2021transformer}, image segmentation \citep{xu2022transformers}, and object detection \citep{islam2022recent}.

Because ViT divides the input image into several patches and needs to model the global relationship directly, researchers focus on how to efficiently use the information of each patch to extract the most distinguishing features of different objects \citep{wu2021rethinking}. The standard ViT introduces positional embedding (PE) to solve this problem, which turned out to be crucial for vision tasks \citep{islam2022recent}. 

Most recent studies change the standard ViT learnable PE, propose a new method to calculate PE, and make the novel PE better express the location information of different patches, the random learnable parameters of PE are still too much to train well \citep{khan2021transformers,su2021roformer,liu2022petr,yang2020position}. Whereas this paper avoids directly changing the PE but instead explores the hidden relationship between the input patches and generates the relationship matrix from a learnable relationship vector. Since the relation embedding (RE) matrix has the same size as the PE, we can add the RE to the original PE or replace it. The advantage is that if we combine RE and PE, we can add more patch-to-patch information on the input without modifying the traditional ViT structure; if we replace PE with RE, the learnable matrix becomes the learnable vector. We compressed the number of learnable parameters from matrices to vectors. Since we played around with the matrix, the effect of adding RE on the training speed is negligible.

\begin{figure}
  \centering
  \includegraphics[width=1\textwidth]{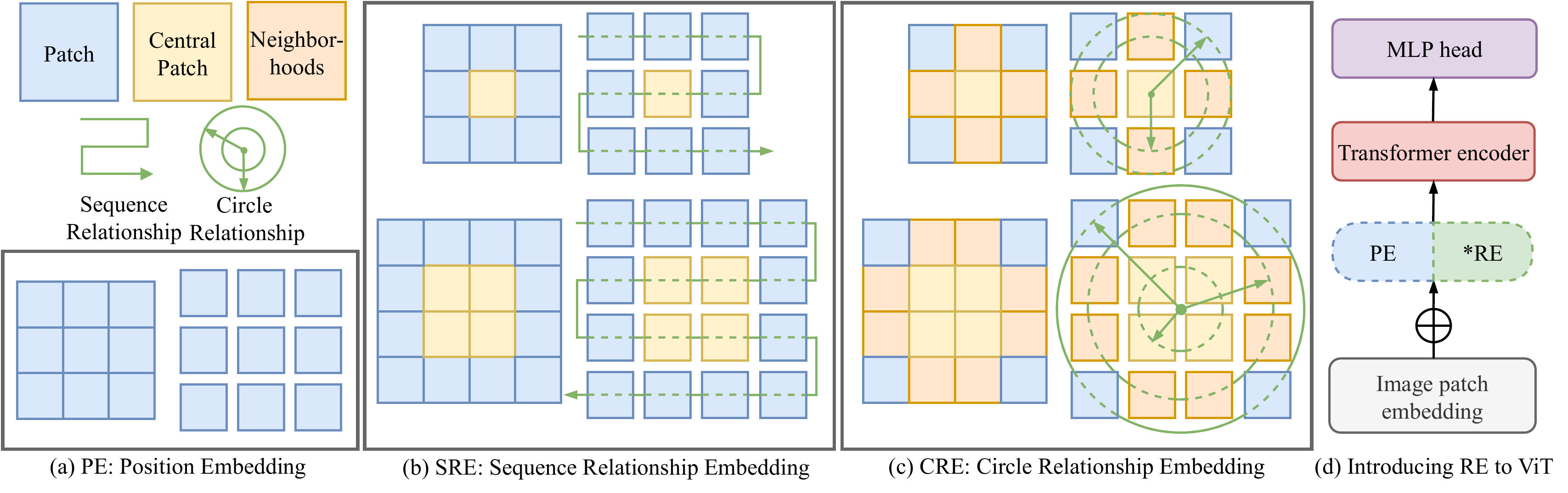}
  \caption{The diagram of different embeddings and the brief structure with the updated ViT. The green line means the relationship between patches. The blue patches represent the image patch, the yellow ones mean the central patch between all patches, and the orange patches are the neighborhoods of the central patch by the 4-neighbors principle. For sequence and circle relationship embedding, we use $3\times3$ patches and $4\times4$ patches as examples to represent the cases where the number of patches is odd or even, respectively. And *RE means the SRE or CRE.}
  \label{fig1}
\end{figure}

Inspired by the sequence \citep{sutskever2014sequence} and the 4-neighbors principle \citep{castleman1996digital} in digital image processing, this paper explores the two possible relationships between patches, the sequence relationship embedding (SRE) and the circle relationship embedding (CRE). In SRE and CRE, we replace the learnable matrix of PE with a vector that encodes, which reduces the two-dimensional random parameters matrix to one dimension. The difference between SRE and CRE is, the former treats the central patch as the center of one sequence, and the patch at the same position on the front and rear sides is the same distance from the central patch. The latter treats the central patch as the center of multiple concentric circles. All patches on the same circle are equidistant from the central patch. Curved arrows and concentric circles can intuitively represent the SRE and the CRE from figure \ref{fig1}. Here we draw on the 4-neighbors principle when calculating CRE, which is also widely used in image erosion \citep{jawas2013image}, edge detection \citep{ziou1998edge}, and other fields \citep{su2009theory}. We assess the SRE and CRE on four public datasets, and the results show that SRE and CRE are effective and provide novel insights into analyzing PE.

\section{Related Work}

ViT treats different patches equally, but in fact, the positional relationship of different patches is not the same, so the introduction of PE helps to improve the performance of ViT. The following shows the four types of PE.

\textbf{Learnable absolute PE.} This PE is proposed by ViT \citep{dosovitskiy2020image} and is also the most widely used PE method. They set a random matrix of fixed dimensions and combined it with patches. The optimizer and ViT parameters are updated synchronously, and the location information of different patches is added to the embeddings.

\textbf{Learnable relative PE.} Swin \citep{liu2021swin} is one of the representatives of learnable relative PE. It uses the relative distance between patches to encode position information, and relative PE uses one-dimensional relative attention. In contrast, the information is encoded in SRE according to the distance between the central patch and other patches.

\textbf{Fixed PE.} Fixed PE uses fixed absolute value coding to represent the positions of different patches. Here \citep{carion2020detr} uses sine and cosine functions for PE and connects the coding information of different frequencies to form the final PE.

\textbf{Other PEs.} In addition to the above three kinds of PE, it also includes PE by using convolutional space invariance \citep{chu2021conditional,wu2021cvt} and using continuous dynamic model \citep{liu2020learning}.

Compared with the above methods, we propose a method of extracting the relationship between patches as a supplement to PE. The size of the RE is consistent with the PE, so we can combine RE with PE to add more information to the latent code or replace PE. Also, the number of learnable parameters of RE is smaller than that of PE.

\section{Method}
Figure \ref{fig1} can generalize the sequence relationship embedding (SRE) and circle relationship embedding (CRE). SRE sets all patches on one sequence, and CRE assumes patches are distributed on circles of different radii. We divide the calculation methods of SRE and CRE into two types according to the parity of the patch. SRE and CRE have the same central patch, and the main difference between SRE and CRE is the distance vector. The details are in the following sections.

\subsection{Sequence Relationship Embedding}
First, considering the PE form, we assume we have $N$ patches and the dimension of PE for each patch is $D$, then $\bm{PE} \in\mathbb{R}^{N\times{D}}$. We can write the learnable position embedding matrix as,
\begin{equation}
    \bm{PE}=\left[\begin{array}{ccc}p_{0,0} & \cdots & p_{0, D-1} \\ \vdots & \ddots & \vdots \\ p_{N-1,0} & \cdots & p_{N-1, D-1}\end{array}\right]_{N \times D}
\end{equation}
where $p_{i,j}$ is the learnable position embedding element $j$ of patch $i$. 

Figure \ref{fig1}(b) illustrates the distance between central patches and others. All patches are ordered as one sequence, and adjacent patches have the same unit distance. We get the SRE of the central patch embedding first and then calculate the distance between the central patch with other patches. The SRE of the central patch is,
\begin{equation}
    \bm{SRC}_{odd}=\left[\begin{array}{lll}p_{\frac{N-1}{2}, 0} & \cdots & p_{\frac{N-1}{2}, D-1}\end{array}\right]_{1 \times D}
\end{equation}
and
\begin{equation}
    \bm{SRC}_{even}=\left[\begin{array}{ccc}p_{\frac{N}{2}-\frac{\sqrt{N}}{2}-1,0} & \cdots & p_{\frac{N}{2}-\frac{\sqrt{N}}{2}-1, D-1} 
                                    \\ p_{\frac{N}{2}-\frac{\sqrt{N}}{2}, 0} & \cdots & p_{\frac{N}{2}-\frac{\sqrt{N}}{2}, D-1} 
                                    \\ p_{\frac{N}{2}+\frac{\sqrt{N}}{2}-1,0} & \cdots & p_{\frac{N}{2}+\frac{\sqrt{N}}{2}-1, D-1} 
                                    \\ p_{\frac{N}{2}+\frac{\sqrt{N}}{2}, 0} & \cdots & p_{\frac{N}{2}+\frac{\sqrt{N}}{2}, D-1}\end{array}\right]_{4 \times D}
\end{equation}
where $\bm{SRC}_{odd}$ and $\bm{SRC}_{even}$ refer to the SRC from an odd number of patches and even number of patches, respectively.

Here we consider $N\geq9$ and have an integer square root. If $N$ is odd, we only have one central patch, whereas we have four central patches if $N$ is even. From Eq.(2) and Eq.(3), we can calculate the index of the central patch based on $N$. Each vector here is learnable and for $\bm{SRC}_{even}$, the $4$ vectors are the same, only the index of the patches are different. So the dimension of the learnable vector could also be $1 \times D$ and broadcast as $4 \times D$.

After getting the SRC, we need to evaluate the distance between the central patch with other patches. Set the distance between adjacent patches as $d$, the distance vector of all patches is,
\begin{equation}
    \bm{dis}_{odd}=[d_0, \cdots, d_{\frac{N-1}{2}-1}, 1, d_{\frac{N-1}{2}+1}, \cdots, d_{N-1}]^T_{N \times 1} \\
\end{equation}
and
\begin{equation}
    \bm{dis}_{even}=[d_0, \cdots, d_{\frac{N}{2}-\frac{\sqrt{N}}{2}-2}, 1, 1, \cdots, 1, 1, d_{\frac{N}{2}+\frac{\sqrt{N}}{2}+1}, \cdots, d_{N-1}]^T_{N \times 1}
\end{equation}

where $d$ can be one hyperparameter or set as an integer to evaluate the distance between adjacent patches. If we set $d=1$ as an example for 9 patches and 16 patches, the distance vector should be $\bm{d}_9=[5,4,3,2,1,2,3,4,5]^T$ and $\bm{d}_{16}=[6,5,4,3,2,1,1,2,2,1,1,2,3,4,5,6]^T$.

When we get the distance vector and SRC, we can calculate the SRE by doing matrix multiplication,
\begin{equation}
    \bm{SRE}_{N \times D}=\bm{dis}_{N \times 1} \times \bm{SRC}_{1 \times D}
\end{equation}
Because of the same dimension, we can simply replace or add SRE to PE. Compare to PE, SRE has fewer learnable parameters and is easy to calculate.

\subsection{Circle Relationship Embedding}
Another relationship between patches is the circle. Figure \ref{fig1}(c) also demonstrates the distance between central patches and others, but the difference is we treat all patches on several concentric circles. All patches on the same circle are the same distance from the central patch because of the same radius. First of all, we need to calculate the CRE for the central patch, which is the same as SRC. The main difference between CRE and SRE is the distance vector.

If the number of patches is odd, we set the center of the circle on the central patch, then consider the 4-neighbors have the same distances from the central patch. We set one of these neighbors as a new central patch, and use the 4-neighbors principle to find the patches that have not yet been traversed. It is considered that all the newly traversed patches are consistent with the initial center patch distance, and the distance is larger than the 4 patches traversed for the first time. We calculate the value of the new distance through the Pythagorean theorem \citep{maor2019pythagorean}. If the index of the central patch is $\frac{N-1}{2}$, the index of 4-neighbors is,
\begin{equation}
    \bm{I}_{neighbors}=\left[\begin{array}{ccc} & \frac{N-1}{2}-\sqrt{N} & 
                        \\ \frac{N-1}{2}-1 & \frac{N-1}{2} & \frac{N-1}{2}+1
                                       \\  & \frac{N-1}{2}+\sqrt{N} & \end{array}\right]
\end{equation}
assume the distance between these 4-neighbors and the center patch is $d$, then we can calculate the next neighbors based on the new four indexes, repeat and set the new distance as $\sqrt{2}d$. So the number of $\bm{dis}$ is $\sum_{i=3,\cdots,\sqrt{N}}\frac{i+1}{2}$. If the number of patches is even, we consider the center of the circle to be in the four central patches, the distance of these patches is 1, and the number of distances is $\sum_{i=4,\cdots,\sqrt{N}}\frac{i}{2}$. The $\bm{dis}$ is,
\begin{equation}
        \bm{dis}_{odd}=[d_{\sum_{i=3,\cdots,\sqrt{N}}\frac{i+1}{2}}, \cdots, d_0, \cdots, d_0, 1, d_0, \cdots, d_0, \cdots, d_{\sum_{i=3,\cdots,\sqrt{N}}\frac{i+1}{2}}]^T_{N \times 1}
\end{equation}
and
\begin{equation}
        \bm{dis}_{even}=[d_{\sum_{i=4,\cdots,\sqrt{N}}\frac{i}{2}}, \cdots, d_0, \cdots, 1, 1, \cdots, 1, 1, \cdots, d_0, d_{\sum_{i=4,\cdots,\sqrt{N}}\frac{i}{2}}]^T_{N \times 1}
\end{equation}
Same as the SRE, $d$ can be the hyperparameter or the fixed number, for the 9 patches and 16 patches from figure \ref{fig1}, the distance vector should be $\bm{d}_9=[2, \sqrt{2},2,\sqrt{2},1,\sqrt{2},2,\sqrt{2},2]^T$ and $\bm{d}_{16}=[2,\sqrt{2},\sqrt{2},2,\sqrt{2},1,1,\sqrt{2},\sqrt{2},1,1,\sqrt{2},2,\sqrt{2},\sqrt{2},2]^T$.
\begin{equation}
\bm{CRC}_{1 \times D}=\bm{SRC}_{1 \times D}
\end{equation}
and
\begin{equation}
\bm{CRE}_{N \times D}=\bm{dis}_{N \times 1} \times {\bm{CRC}_{1 \times D}}
\end{equation}
because the CRE of the central patch (CRC)is equal to SRC, we can get the CRE by doing matrix multiplication. From figure \ref{fig1}(d), we could combine PE with SRE or CRE, even replace PE.

\section{Experiment}
We choose four popular datasets (CIFAR10, CIFAR100 \citep{krizhevsky2009learning}, Tiny-ImageNet \citep{ILSVRC15}, and Flowers102 \citep{Nilsback08}) to evaluate our method, and trained the network for 400 epochs. The batch size was 128, and we use AdamW \citep{loshchilov2017decoupled} as the optimizer. The learning rate is 0.001 and combines the cosine learning rate decay \citep{loshchilov2016sgdr}. Our experiments use these smaller datasets and smaller batch sizes. Therefore, we were able to train our models using one GPU like Nvidia GeForce Titan X. Each experiment is performed five times, and the average value is taken as the output.

\begin{table}
  \renewcommand\arraystretch{1.2}
  \setlength{\tabcolsep}{12pt}
  \caption{Top-1 accuracy comparison on different datasets}
  \centering
    \begin{tabular}{lcccc}
    \toprule
    \textbf{Method} & \multicolumn{1}{l}{\textbf{CIFAR10}} & \multicolumn{1}{l}{\textbf{CIFAR100}} & \multicolumn{1}{l}{\textbf{T-ImageNet}} & \multicolumn{1}{l}{\textbf{Flowers102}} \\ \hline
    \multicolumn{5}{c}{ViT-Base}                                                                                                                                                      \\ \hline
    PE             & 95.68                                & 76.25                                 & 54.86                                   & 85.64                                   \\ 
    SRE            & 96.15(+0.47)                         & 76.74(+0.49)                          & 55.14(+0.28)                            & 86.23(+0.59)                            \\
    SRE+PE         & 96.14(+0.46)                         & 77.08(+0.83)                          & 55.38(+0.52)                            & 86.76(+1.12)                            \\ 
    CRE            & 96.19(+0.51)                         & 77.06(+0.81)                          & 55.29(+0.43)                            & 86.48(+0.84)                            \\
    CRE+PE         & \textbf{96.28(+0.60)}                         & \textbf{77.35(+1.10)}                          & \textbf{55.45(+0.59)}                            & \textbf{86.83(+1.19)}                            \\ \hline
    \multicolumn{5}{c}{T2T-Base}                                                                                                                                                      \\ \hline
    PE             & 96.78                                & 79.29                                 & 57.56                                   & 88.13                                   \\ 
    SRE            & 96.93(+0.15)                         & 79.71(+0.42)                          & 57.77(+0.21)                            & 88.09(-0.04)                            \\
    SRE+PE         & 97.32(+0.54)                         & 79.88(+0.59)                          & 57.92(+0.36)                            & 88.34(+0.21)                            \\ 
    CRE            & 96.98(+0.20)                         & 79.78(+0.49)                          & 57.89(+0.33)                            & 88.24(+0.11)                            \\
    CRE+PE         & \textbf{97.45(+0.67)}                         & \textbf{80.23(+0.94)}                          & \textbf{58.56(+1.00)}                            & \textbf{88.59(+0.46)}                            \\ \hline
    \multicolumn{5}{c}{Swin-Base}                                                                                                                                                     \\ \hline
    PE             & 96.82                                & 79.78                                 & 58.69                                   & 89.46                                   \\ 
    SRE            & 96.73(-0.09)                         & 80.11(+0.33)                          & 58.97(+0.28)                            & 89.58(+0.12)                            \\
    SRE+PE         & 97.39(+0.57)                         & 80.33(+0.55)                          & 59.45(+0.76)                            & 89.76(+0.30)                            \\ 
    CRE            & 97.20(+0.38)                         & 80.19(+0.41)                          & 59.23(+0.54)                            & 89.69(+0.23)                            \\
    CRE+PE         & \textbf{97.74(+0.92)}                         & \textbf{80.45(+0.67)}                          & \textbf{59.78(+1.09)}                            & \textbf{89.91(+0.45)}                            \\ \bottomrule
    \end{tabular}
\end{table}

To evaluate the robustness of SRE and CRE, we implement them on three different ViT structures. Here ViT-Base, T2T-Base, and Swin-Base mean the ViT-B/16 \citep{dosovitskiy2020image}, T2T-ViT-14 \citep{yuan2021tokens}, and Swin-T \citep{liu2021swin}. PE is the original position embedding method from these basic ViT models. We treat PE as the baseline to compare our methods. From the table, we can summarize several points.

1) SRE and CRE can directly replace or combine with PE in three ViT backbones, because of the same matrix dimension we discussed in the method part.

2) We observe an improved ViT performance on these datasets. We hypothesize that it comes from the additional inductive bias of SRE or CRE, which are more focused on patch-to-patch information, while PE only pays attention to the location information of different patches in space.

3) The combination of both consistently outperform just using PE or these relationship embeddings. By combining PE with SRE or CRE, the latent input code of patches would be more suitable for the ViT structure. We presume SRE and CRE can be supplementary to PE.

4) Using CRE or combined PE with CRE yields better results than SRE. Image is 2-D data, and if it omits the color information, it should be the same weight from the center pattern to the surroundings from the intuition. So the relation will be more precise if we treat the surroundings as a circle rather than a sequence. So we conclude CRE is a better relationship to describe the distance between patches.

\section{Conclusion}
In this work, we introduce sequence relationship embedding (SRE) and circle relationship embedding (CRE) to edit the relationship between patches and the central patch. We use the distance vector times one learnable vector to construct SRE or CRE, reducing the random parameters in position embedding (PE) from two dimensions to one dimension. Both SRE and CRE can replace or combine with PE. When combined, it outperforms just using PE. However, these two relationships still need analysis from more perspectives like training speed, latent representation, and visualization and also need more ViT backbones and datasets to evaluate the performance. Exploring these issues will bring us closer to finding more perspectives on studying PE and ViT.


\bibliographystyle{svrhm_2022}
\bibliography{svrhm_2022}

\end{document}